\pdfoutput=1

\documentclass[11pt]{article}

\usepackage[final]{acl}

\usepackage{times}
\usepackage{latexsym}

\usepackage[T1]{fontenc}

\usepackage[utf8]{inputenc}

\usepackage{microtype}

\usepackage{inconsolata}

\usepackage{graphicx}

\title{Breaking the Ceiling of the LLM Community by Treating Token Generation as a Classification for Ensembling}

\usepackage{authblk}
\author{
 \normalsize\textbf{Yao-Ching Yu\textsuperscript{1}},
 \textbf{Chun-Chih Kuo\textsuperscript{1}},
 \textbf{Ziqi Ye\textsuperscript{2}},
 \textbf{Yu-Cheng Chang\textsuperscript{1}},
 \textbf{Yueh-Se Li\textsuperscript{1}}
\\
 \normalsize\textsuperscript{1}AI Lab, TrendMicro~
 \textsuperscript{2}Shanghai Jiao Tong University
\\
 \normalsize{\texttt{\{yaoching\_yu,tony\_kuo,ryker\_chang,joseph\_li\}@trendmicro.com}, \texttt{yzq\_9907@sjtu.edu.cn}}
}

\usepackage{amsmath}
\usepackage{amssymb}
\usepackage{booktabs}
\usepackage{array}
\usepackage{colortbl} 
\usepackage{xcolor} 
\usepackage{multirow} 

\newcommand{\tokenSpace}{\rule{0.5em}{0.125em}}

\definecolor{beaublue}{rgb}{0.74, 0.83, 0.9}
\newcommand{\tightcellcolor}[2]{\setlength{\fboxsep}{2pt}\colorbox{#1}{#2}}

\begin{document}
\maketitle
\begin{abstract}
Ensembling multiple models has always been an effective approach to push the limits of existing performance and is widely used in classification tasks by simply averaging the classification probability vectors from multiple classifiers to achieve better accuracy. However, in the thriving open-source Large Language Model (LLM) community, ensembling methods are rare and typically limited to ensembling the full-text outputs of LLMs, such as selecting the best output using a ranker, which leads to underutilization of token-level probability information. In this paper, we treat the \textbf{G}eneration of each token by LLMs \textbf{a}s a \textbf{C}lassification (\textsc{\textbf{GaC}}) for ensembling. This approach fully exploits the probability information at each generation step and better prevents LLMs from producing early incorrect tokens that lead to snowballing errors. In experiments, we ensemble state-of-the-art LLMs on several benchmarks, including exams, mathematics and reasoning, and observe that our method breaks the existing community performance ceiling. Furthermore, we observed that most of the tokens in the answer are simple and do not affect the correctness of the final answer. Therefore, we also experimented with ensembling only key tokens, and the results showed better performance with lower latency across benchmarks.\footnote{Our code: \href{https://github.com/yaoching0/GaC}{https://github.com/yaoching0/GaC}}
\end{abstract}

\section{Introduction}
\label{Introduction}
Large Language Models (LLMs) have demonstrated remarkable capabilities in a wide range of natural language processing tasks \citep{achiam2023gpt,touvron2023llama}. Over time, new and more powerful LLMs are continually being released, pushing the boundaries of the LLM community \citep{meta2024llama3,qwen2blog}. Due to the diversity of data sources, architectures and training methods, different LLMs have strengths and weaknesses in different tasks and contexts \citep{jiang2023llm}. In addition to investing significant resources in training a superior LLM, ensembling multiple existing models is another effective way to break through the community performance ceiling \citep{huang2016snapshot}, especially given the current trend in the open source LLM community to contribute only model weights rather than training data and procedures \citep{allenai2023olmo}.

\begin{figure}[t]
  \centering
  \includegraphics[width=\columnwidth]{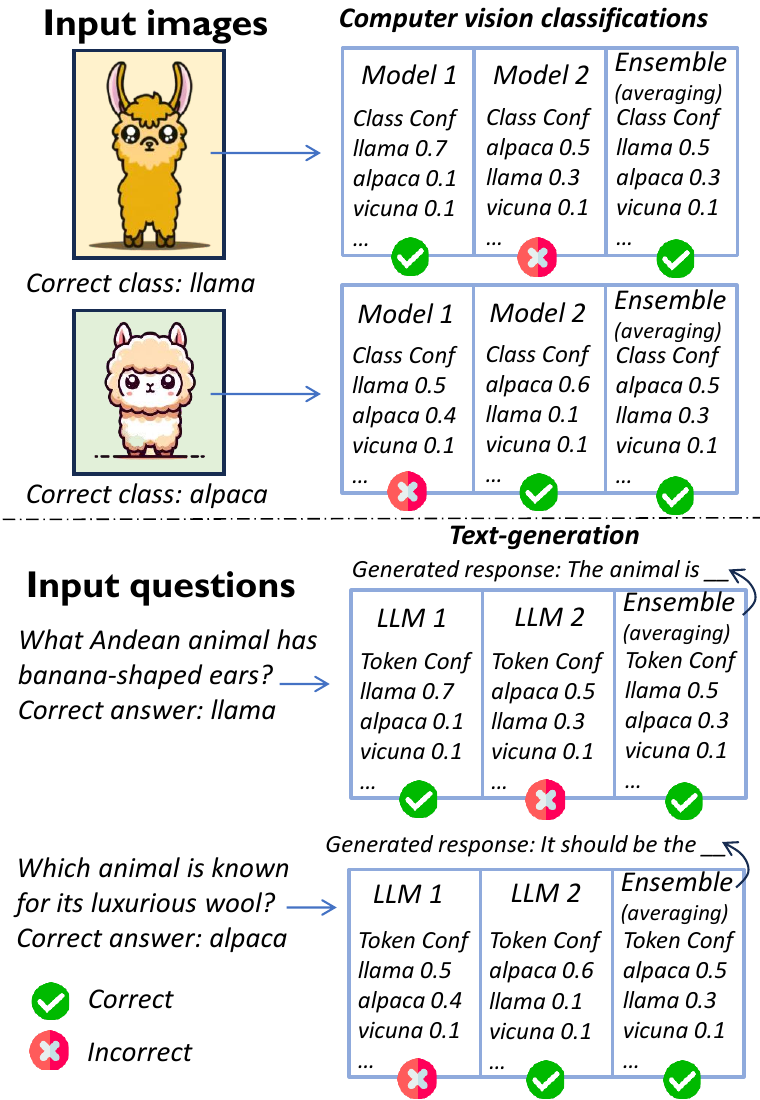}
  \caption{Motivation of \textsc{GaC}. The upper part shows CV classification ensemble, while the lower part illustrates ensemble at one text generation step.}
  \label{fig_1}
\end{figure}

Taking computer vision (CV) classification as an example, it is common to ensemble the output probability vectors of multiple models (e.g. by averaging) to achieve superior results \citep{krizhevsky2017imagenet}. This approach remains effective even with recent CV models. As shown in Tab.\ref{tab1}, we selected several common CV models \citep{chen2022repghost, tan2019efficientnet,wang2021pvtv2} for ensembling and observed better accuracy on ImageNet \citep{deng2009imagenet} compared to using a single model. Similarly, the popular decoder-only LLM architecture generates text by producing tokens one by one, with each generation step resulting in a probability vector of the length of the vocabulary. Inspired by CV, we propose to treat each \textbf{G}eneration step \textbf{a}s a \textbf{C}lassification task (\textsc{\textbf{GaC}}), and by ensembling multiple models, we can achieve higher accuracy, as shown in Fig.\ref{fig_1}. There is already work that simplifies problems into binary tasks, exploiting the collective wisdom of LLMs and achieving better results \citep{schoenegger2024wisdom}, demonstrating the feasibility of this approach.

\begin{table}[t]
    \centering
    \footnotesize
    \setlength{\tabcolsep}{2mm}{
    \begin{tabular}{lcc}
        \toprule[1.2pt]
        \textbf{CV Models on ImageNet} & \emph{Acc} [\%] & \emph{ECE} \\
        \midrule
        EfficientNet-B1 & 78.55 & 0.072 \\
        RepGhostNet & 78.81 & 0.053 \\
        PVTv2-B1 & 78.71 & 0.119 \\
        \midrule
        \multicolumn{3}{c}{\emph{Ensembled CV Models (Averaged)}} \\
        \midrule
        EfficientNet-B1 + PVTv2-B1 & 80.20\textbf{$\uparrow$1.49} & - \\
        EfficientNet-B1 + RepGhostNet & 80.06\textbf{$\uparrow$1.25} & - \\
        EfficientNet-B1 + RepGhostNet & & \\
        \multicolumn{1}{r}{+ PVTv2-B1} & 80.62\textbf{$\uparrow$1.81} & - \\
        \midrule
        \textbf{LLMs on MMLU} & \emph{Acc} [\%] & \emph{ECE}  \\
        \midrule
        Llama-3-70b-Instruct & 79.68 & 0.095 \\
        Qwen1.5-72b-chat & 77.79 & 0.089 \\
        Yi-34B-Chat & 72.75 & 0.090 \\
        \bottomrule[1.2pt]
    \end{tabular}}
    \caption{Performance of various CV models on ImageNet and LLMs on MMLU. \textbf{$\uparrow$} indicates improvement over a single model.}
    \label{tab1}
\end{table}

Another advantage is that early errors in LLMs often snowball into later errors \citep{zhang2023language}. Ensembling during generation helps prevent the generation of inaccurate tokens at each step, thereby reducing misleading cues for subsequent token generation. In this paper, we conducted experiments at several points in time between November 2023 and June 2024, ensembling available state-of-the-art (SOTA) LLMs up to each of these points. We found that this approach \textbf{significantly} outperformed any single model available at those times as well as existing methods on five popular benchmarks involving subject examination, mathematics, reasoning, and knowledge-based QA.

In addition, we found that for text generation it seemed unnecessary to ensemble at every step. For example, for the question "\textit{What Andean animal has banana-shaped ears?}" shown in Fig.\ref{fig_1}, the most critical part is for the LLM to generate the key token "\textit{llama}". The initial part of the answer "\textit{It should be \_}" or "\textit{The animal is \_}" do not significantly affect the correctness of the final answer. Ideally, the step that produces the token "\textit{llama}" is the one we want to ensemble.
\label{ensemble-with-t}

Studies in CV classification have also shown that most samples are "simple" and can be correctly classified by most models \citep{wang2017idk}, including cost-efficient ones, making the use of expensive models wasteful. To address this, CV classification used cascade inference \cite{jazbec2024towards,enomoro2021learning}, where a gate model passes a sample to a more powerful model only if its confidence falls below a threshold, thereby improving efficiency. Obviously, it is very important for cascading that the confidence of the gate model accurately reflects the accuracy. To ensure that LLMs are also suitable as gate models, we measured the Expected Calibration Error (ECE) \citep{guo2017calibration} of CV models and LLMs on ImageNet and MMLU \citep{hendrycks2020measuring}, as shown in Tab.\ref{tab1}. ECE is a metric that reflects the difference between a model's confidence and its accuracy. We found that the ECE of CV models and LLMs were close. Therefore, in this paper, we also applied the cascade inference to LLMs by ensembling only the "key" tokens to speed up generation. Our experiments showed that this approach consistently achieved better performance with lower latency across different benchmarks.

In summary, we propose a framework \textsc{GaC}, which allows multiple heterogeneous LLMs to jointly decide the next generated token during text generation:

\begin{itemize}
    \item When we ensemble at every generation step, experiments show that it outperforms any single SOTA LLM from different time periods and existing ensemble methods.
    \item By using cascade to ensemble only at important steps during text generation, experiments show better results with lower latency.
\end{itemize}

\section{Analysis and Prior Work}
In this chapter, we review previous LLM ensemble studies and their features, as well as the problems our approach addresses. Previous studies can be categorized as follows:

\textbf{Output-level ensemble} methods select multiple candidate models and use their complete outputs for ensembling. \citet{jiang2023llm} trained an additional ranking model (\emph{PairRanker}) to score each candidate output and select the best one. \citet{lu2023routing} and \citet{shnitzer2023large} trained a router to select the most appropriate candidate model given a question. However, these methods are limited to the existing candidate outputs and become ineffective if all the outputs are incorrect. Other studies have trained a fusion model to blend the outputs \citep{jiang2023llm,wang2023fusing}, overcoming the limitation of selecting only existing candidate outputs and often achieving superior results. However, the generalization of the fusion model is a major challenge, and they cannot fully exploit the probability information from each generation step.

\textbf{Weight-level ensemble} methods merge the weights of multiple models and are primarily used in multi-task learning \citep{yadav2024ties}. The expectation is that the merged model will inherit capabilities across multiple tasks. However, a limitation is that the architectures of the models to be merged must be homologous, which limits the use of the capabilities of the LLM community. And it is rare to observe that the merged model outperforms the original models \citep{yu2023language}.

\textbf{Training-level ensemble} like FuseLLM \citep{wan2024knowledge} uses the output probability vectors of multiple models during training to ensemble as labels, rather than one-hot labels. In effect, this is a specific form of distillation that allows the model being trained to gain more information from the probability outputs of the ensembled (teacher) models. However, distillation is mainly used to improve small models, making it difficult to further improve the SOTA LLMs.

\textbf{Our work} can overcome the above limitations by ensembling at each generation step, allowing the output not to be confined to the original candidate output space and homologous architectures, while fully exploiting the probability information at each step. Our experiments in Sec.\ref{sec4} will also show that the ensemble consistently outperforms any single model, even SOTA LLMs. The main challenge, however, is that different LLMs typically have inconsistent vocabularies, leading to different dimensions in the probability vectors produced by different models. The most intuitive solution is to take the union of the vocabularies of the ensembled LLMs, denoted $V^{U}$, which includes all tokens from the participating models. Then, at each generation step, the output is first mapped to this union space $\mathbb{R}^{|V^U|}$ before ensembling.

\begin{figure}[t]
  \centering
  \includegraphics[width=\columnwidth]{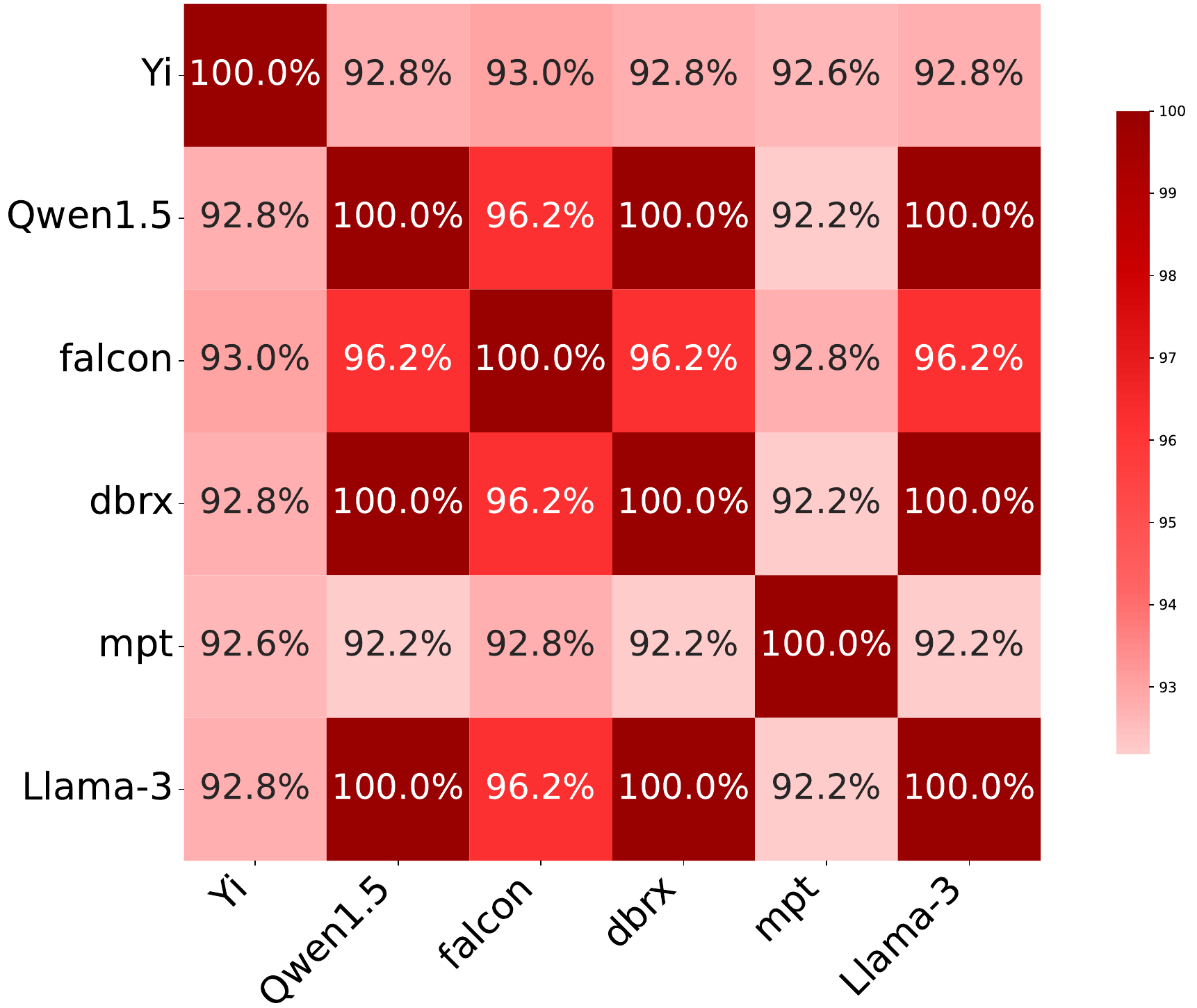}
  \caption{The rate of identical tokenization for Oxford 5000 common words between different LLMs.}
  \label{fig_2}
\end{figure}

A potential problem with this approach is that different models may tokenize the same word differently, leading to conflicts. However, most mainstream LLMs use BPE or BBPE \citep{sennrich2015neural,wang2020neural} to train tokenizers on sampled corpora, which tend to have similar sources (e.g. CommonCrawl) and distributions. This results in consistent tokenization for common words. For example, both Qwen1.5 and Llama3 \citep{bai2023qwen,meta2024llama3} tokenize the word "\emph{ alphabetically}" into ["\emph{Ġalphabet}", "\emph{ically}"]. If both models intend to output this word, they will assign a higher probability to "\emph{Ġalphabet}" first. We selected several popular LLMs \citep{young2024yi,databricks2023dbrx,almazrouei2023falcon} and tokenized 5,000 commonly used English words \citep{oxford5000}, and then calculated the proportion of identical tokenization results between each pair of LLMs, as shown in Fig.\ref{fig_2}. The proportion is above $90\%$ for all pairs, indicating that such conflicts can be ignored in most cases.


\section{Proposed Method}
In this section, we will first introduce the overall ensemble process of our \textsc{GaC} framework, and then explain the details in the following subsections.

\begin{figure*}[ht]
  \centering
  \includegraphics[width=\textwidth]{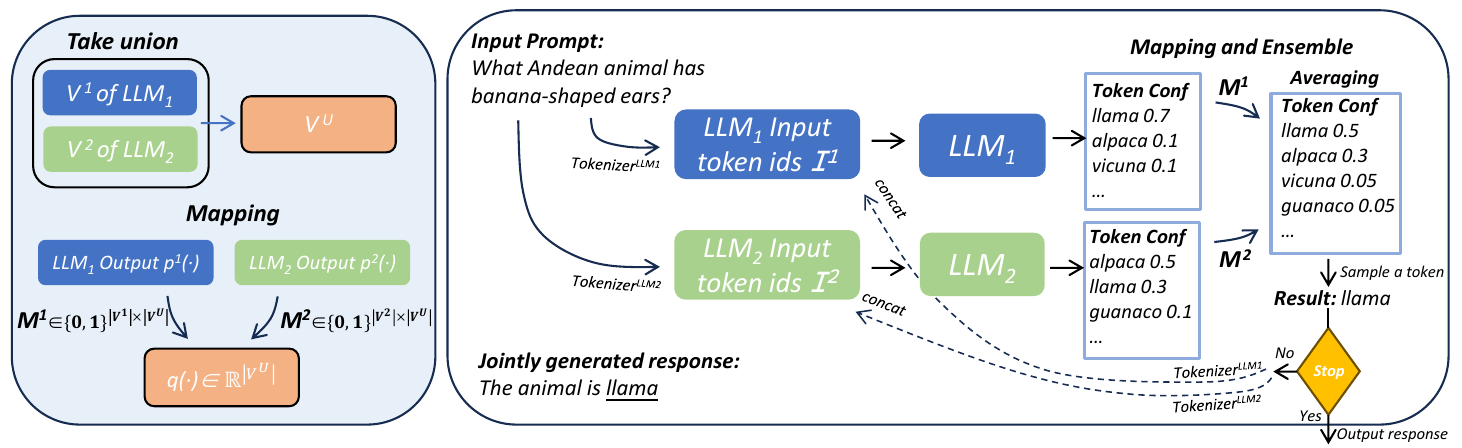}
   \caption{Overview of \textsc{GaC}. The left side shows the creation of the mapping matrix, and the right side shows the ensembling during text generation with two LLMs.}
  \label{fig_3}
\end{figure*}

\subsection{Overall Process of \textsc{GaC}}
\label{sec3-1}
When generating text, LLMs output a probability vector of the same dimension as their vocabulary. Given \(n\) LLMs to be ensembled, we first take the union of their vocabularies and create a mapping matrix that can project the probability vectors to the union dimensions (Sec.\ref{sec3-2}). At each generation step, all LLMs produce outputs that are mapped to the union vocabulary dimensions and ensembled to sample the next token. The tokenizer of each LLM then converts the sampled token into token IDs for the next step (Sec.\ref{sec3-3}). As mentioned in Sec.\ref{ensemble-with-t}, not all generation steps have the necessity for ensembling, so we also try to ensemble only certain key tokens (Sec.\ref{sec3-4}).

\subsection{Creating the Union Mapping}
\label{sec3-2}
Given \(\{\textit{LLM}_1, \textit{LLM}_2, \ldots, \textit{LLM}_n\}\) to ensemble, with their respective vocabularies \(\{V^1, V^2, \ldots,\allowbreak V^n\}\), we first take the union of the vocabularies:
\begin{equation}
\label{eq1}
V^{U} = \bigcup_{i=1}^{n} V^i.
\end{equation}
During this process, we record the positions of tokens from \(V^i\) in \(V^{U}\) and create corresponding mapping matrices \(\mathbf{M}^{i} \in \{0,1\}^{|V^i| \times |V^{U}|}\).

\subsection{\textsc{GaC} Ensembling}
\label{sec3-3}
At the start of text generation, we convert the input \textit{prompt} into token ID sequences for each LLM. We denote the tokenizer of \(\textit{LLM}_i\) as \(\mathcal{T}^{\,i} : \textit{text} \rightarrow (\tau_1, \tau_2, \ldots, \tau_m)\), which converts the input text into a sequence of token IDs. We calculate:
\begin{equation}
\label{eq2}
\mathcal{I}^{\,i} = \mathcal{T}^{\,i}(\textit{prompt}) \quad \text{for } i = 1, \ldots, n
\end{equation}
where \(\mathcal{I}^{\,i}\) is the input token ID sequence for \(\textit{LLM}_i\).

For each generation step, we input \(\mathcal{I}^{\,i}\) into \(\textit{LLM}_i\) to obtain \(p^i(\cdot\,|\,\mathcal{I}^{\,i}) \in \mathbb{R}^{|V^i|}\), which represents the probability vector for the next token. These vectors are then mapped to the union vocabulary dimensions and averaged:
\begin{equation}
\label{eq3}
q(\cdot) = \frac{1}{n} \sum_{i=1}^{n} p^i(\cdot\,|\,\mathcal{I}^{\,i}) \cdot \mathbf{M}^{i},
\end{equation}
where \(q(\cdot)\) is the ensemble probability vector. In Sec.\ref{sec4-3}, we experimented with different ensemble weights and decided to use the average. We then sample a token \(x \sim q(\cdot)\) as the result of this step. Finally, the sampled token is converted back into token IDs for each LLM and appended to \(\mathcal{I}^{\,i}\):
\begin{equation}
\label{eq4}
\mathcal{I}^{\,i} \leftarrow \mathcal{I}^{\,i}{}^{\frown} \mathcal{T}^{\,i}(x) \quad \text{for } i = 1, \ldots, n
\end{equation}

We repeat (\ref{eq3}) and (\ref{eq4}) until the stopping criteria are met, such as outputting an end-of-sentence token or reaching the maximum length, as shown in Fig.\ref{fig_3}.\footnote{\footnotesize We provide step-by-step examples in Appx.\ref{sec:appendix-example}.} In our implementation, different LLMs run in parallel on different GPUs, so the duration of each step is equal to the time taken by the slowest LLM. Since we have not modified a complete forward pass, our approach is compatible with techniques such as vLLM, DeepSpeed, quantization, and hardware optimizations \citep{kwon2023efficient, rasley2020deepspeed}.

\subsection{Ensembling Key Tokens with Threshold}
\label{sec3-4}
As mentioned in the last part of Sec.\ref{ensemble-with-t}, most tokens do not significantly affect the correctness of the response. From Tab.\ref{tab1}, we can see that LLMs and CV models have similar ECE levels, suggesting that the  confidence scores of LLMs may reflect accuracy to some extent. Therefore, we also experiment with ensembling only the steps with confidence below a threshold \( t \). We choose a model as the gate, denoted \(\textit{LLM}_g\), and use its maximum probability at each step as the confidence score. During the ensemble, we replace the original (\ref{eq3}) with:
\begin{equation}
\label{eq5}
\resizebox{.93\hsize}{!}{$
q(\cdot) =
\begin{cases}
\frac{1}{n} \sum_{i} p^i(\cdot\,|\,\mathcal{I}^{\,i}) \cdot \mathbf{M}^{i} & \textit{if } \max(p^g(\cdot\,|\,\mathcal{I}^{\,g})) \leq t \\
p^g(\cdot\,|\,\mathcal{I}^{\,g}) \cdot \mathbf{M}^{g} & \textit{otherwise}.
\end{cases}
$}
\end{equation}

Note that apart from \(\textit{LLM}_g\), the other LLMs are not computed at every step, so their KV caches become stale. While there has been research using partial KV caches \citep{barad2023leveraging}, for simplicity our work disables the KV caches of all LLMs except \(\textit{LLM}_g\). This is an area for improvement and is listed in our future work.

\section{Experiments}
\label{sec4}
\subsection{Overview}
\label{sec4-1}

In this section, we first present the experimental setup, including the benchmarks and hardware used (Sec.\ref{sec4-2}). We then test the effects of different ensemble parameters for \textsc{GaC} (Sec.\ref{sec4-3}) and compare it with other methods (Sec.\ref{sec4-4}). We also select SOTA LLMs available at different times for ensembling to explore the performance ceiling at each time period (Sec.\ref{sec4-5}). Finally, we experiment with thresholded ensembling to explore variations in latency and performance (Sec.\ref{sec4-6}).

\subsection{Experimental Settings}
\label{sec4-2}

\textbf{Benchmarks.} \textsc{GaC} is not limited to specific tasks, so we tested it as broadly as possible. We selected a total of five benchmarks. For general capabilities, we chose MMLU \citep{hendrycks2020measuring}. For maths, we utilized GSM8K \citep{cobbe2021training}. For reasoning, we employed BBH \citep{suzgun2023challenging}. For knowledge capabilities, we included TriviaQA \citep{joshi2017triviaqa} and NaturalQuestions (NQ) \citep{kwiatkowski2019natural}. Note that all scores, including those for individual models, were computed locally under the same environment to ensure fairness\footnote{\footnotesize Please see  Appx.\ref{sec:appendixA} for more benchmarks details.}, using lm-evaluation-harness\footnote{\footnotesize \href{https://github.com/EleutherAI/lm-evaluation-harness}{https://github.com/EleutherAI/lm-evaluation-harness}} v0.4.1 \cite{eval-harness}.

\vspace{0.5\baselineskip}
\noindent\textbf{Hardware and Latency.} Each LLM was loaded on 1(n) A100 GPU(s) according to its memory requirements\footnote{\footnotesize We listed each model and its hardware in Appx.\ref{sec:appendixB}.}, using naive model parallelism \citep{huggingface2024} without optimization for inference. During ensembling, different LLMs were loaded on separate GPU(s) and executed in parallel, managed and communicated via Ray \citep{moritz2018ray}. We also recorded the latency (ms/token). Each model performs a "dry run" after being loaded onto the GPU, generating 1024 tokens to warm up CUDA before experimentation, following the practice of \citet{mehta2024openelm}.

\subsection{Exploring Ensemble Parameters}
\label{sec4-3}

\begin{table}[t]
\footnotesize
\centering
\setlength{\tabcolsep}{2pt}
\begin{tabular}{>{\raggedright\arraybackslash}p{5cm}cc} 
    \toprule
    \textbf{Model} & \textit{Acc} [\%] & \textit{ECE} \\
    \midrule
    OpenChat-3.5-0106 & 64.53 & 0.0833 \\
    Qwen1.5-14B-Chat & 67.20 & 0.1312 \\
    SOLAR-10.7B-Instruct-v1.0 & 64.48 & 0.2884 \\
    \midrule
    Yi-34B-Chat & 72.75 & 0.0903 \\
    Qwen1.5-32B-Chat & 75.12 & 0.1003 \\
    Nous-Hermes-2-Mixtral-8x7B-DPO & 72.65 & 0.0789 \\
    \bottomrule
\end{tabular}
\caption{Two sets of LLMs of different sizes on MMLU. Smaller models on top, larger models on bottom.}
\label{table2}
\end{table}

\begin{figure}[t]
  \centering
  \includegraphics[width=\columnwidth]{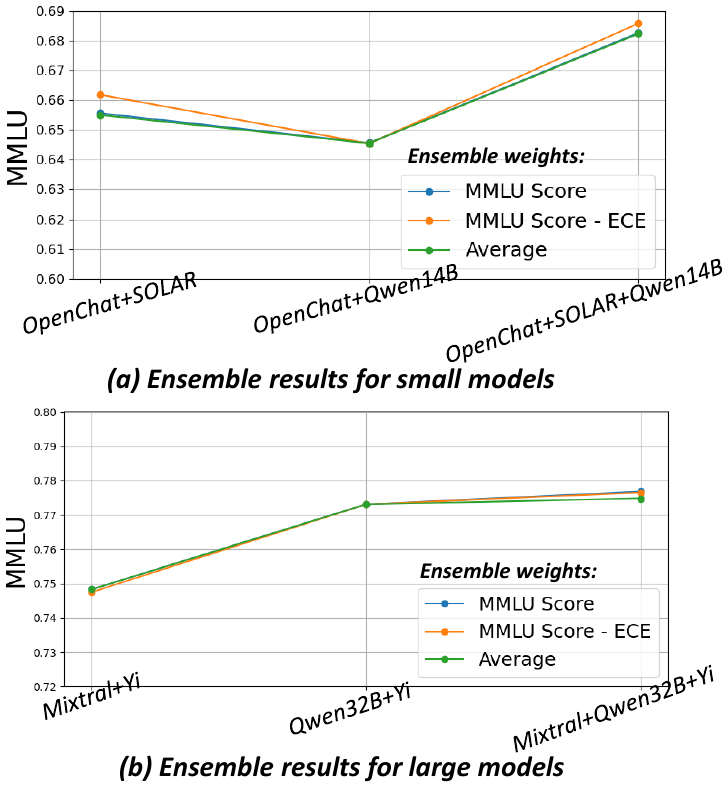}
  \caption{Results of \textsc{GaC} ensemble with different weights for models from Tab.\ref{table2}. Smaller models ensembles on top, larger ones on bottom. The x-axis shows names participating in the ensemble (abbreviated).}
  \label{fig_4}
\end{figure}

\textbf{Different Ensemble Weights.} Before proceeding with further experiments, we tested different ensemble weights for \textsc{GaC}. We used a simple averaging of the probabilities from each model in Eq.\ref{eq3}. We now replace Eq.\ref{eq3} by \(\frac{1}{\sum_{i}w^{i}} \sum_{i}w^{i} p^i(\cdot\,|\,\mathcal{I}^{\,i}) \mathbf{M}^{i}\), where \(w^{i}\) is the ensemble weight for \(\textit{LLM}_{i}\). We set \(w^{i}\) separately to each LLM's score on MMLU, the MMLU score minus the ECE and 1 (i.e. averaging). We selected two sets of LLMs \citep{wang2023openchat,kim2023solar,young2024yi,bai2023qwen,nousmixtral} listed in Tab.\ref{table2} with different model sizes for \textsc{GaC} ensemble. The results on MMLU are shown in Fig.\ref{fig_4}. We observed no significant differences between the different weights, so for simplicity we decided to use averaging.

\begin{table}[t]
\footnotesize
    \centering
    \begin{tabular}{@{}lcc@{}}
        \toprule
        \textbf{Model} &  \textit{Acc} [\%] & $\Delta$ \\
        \midrule
        Qwen1.5-14B-Chat & 67.20 & - \\
        Phi-3-mini-4k-instruct & 67.11 & - \\
        OpenChat-3.5-0106 & 64.53 & - \\
        SOLAR-10.7B-Instruct-v1.0 & 64.48 & - \\
        \midrule
        \textsc{GaC} (Qwen1.5 + Phi-3) & 69.91 & +2.71 \\
        \textsc{GaC} (Qwen1.5 + Phi-3 + OpenChat) & 70.40 & +3.20 \\
        \textsc{GaC} (Qwen1.5 + Phi-3 + OpenChat & & \\
        \multicolumn{1}{r}{+ SOLAR)} & 70.01 & +2.81 \\
        \bottomrule
    \end{tabular}
    \caption{Performance of different ensemble sizes on MMLU. The top part lists individual models, while the bottom part shows ensemble results (model names abbreviated). $\Delta$ represents the improvement compared to the best individual model.}
    \label{table:x}
\end{table}

\vspace{0.5\baselineskip}
\noindent\textbf{Different Ensemble Sizes.} We also investigated the effect of ensembling different numbers of models on performance. Specifically, we selected four popular small LLMs \citep{bai2023qwen,kim2023solar,wang2023openchat,abdin2024phi}. After sorting the models based on their MMLU scores from highest to lowest, we sequentially ensembled the top 2, 3, and 4 models. As shown in Tab.\ref{table:x}, the results indicate that the performance gain from ensembling does not always increase with the addition of more models. The highest score was achieved when ensembling three models, while ensembling all four models resulted in a slight decrease in performance (from 70.40 to 70.01).

\subsection{Comparison with Other Methods}
\label{sec4-4}
\begin{table*}[t]
\centering
\footnotesize
\setlength{\tabcolsep}{3pt}
\begin{tabular}{>{\raggedright\arraybackslash}p{0.5cm}>{\raggedright\arraybackslash}p{4.2cm}ccccccc}
    \toprule
    \textbf{Id} & \textbf{Models} & \textbf{MMLU} & \textbf{GSM8K} & \textbf{BBH} & \textbf{TriviaQA} & \textbf{NQ} & \textbf{Avg.} & \textbf{Latency} \\
    \midrule
    1 & openchat\_3.5 & 63.87 & 68.46 & 47.96 & 68.12 & 29.75 & 55.63 & 28.01\scriptsize\textit{ms/token} \\
    2 & Nous-Hermes-2-SOLAR-10.7B & 64.88 & 72.86 & 49.92 & 71.33 & 32.21 & 58.24 & 50.89\scriptsize\textit{ms/token} \\
    \midrule
    \multicolumn{9}{c}{\emph{Ensemble Results for the Above Two Models (openchat and SOLAR)}} \\
    \midrule
    3 & {LLM Blender (PairRanker)} & 64.23 & 74.07 & 50.18 & 70.55 & 32.55 & 58.32 & 57.86\scriptsize\textit{ms/token} \\
    4 & {OAssistRM} & 64.87 & 72.93 & 49.02 & 70.06 & 31.64 & 57.70 & 55.62\scriptsize\textit{ms/token} \\
    5 & {UltraRM} & 64.94 & \tightcellcolor{beaublue}{75.51} & 50.65 & 71.23 & 32.03 & 58.87 & 74.51\scriptsize\textit{ms/token} \\
    6 & \textsc{GaC} (\textit{\textbf{ours}}) & \tightcellcolor{beaublue}{66.51} & 74.30 & \tightcellcolor{beaublue}{51.19} & \tightcellcolor{beaublue}{72.50} & \tightcellcolor{beaublue}{33.82} & \tightcellcolor{beaublue}{59.66} & 51.32\scriptsize\textit{ms/token} \\
    \midrule[0.1pt]
    7 & {FuseLLM} & 63.94 & 65.50 & 46.32 & 64.57 & 29.06 & 53.88 & 28.31\scriptsize\textit{ms/token} \\
    8 & \textsc{GaC}$^{t=0.5}_{7.68\%}$ (\textit{\textbf{ours}}) & 65.14 & 73.18 & 50.32 & 69.68 & 31.75 & 58.01 & 31.34\scriptsize\textit{ms/token} \\
    \midrule[1pt]
    9 & Mixtral-8x7B-Instruct-v0.1 & 70.89 & 66.82 & 49.84 & 76.54 & 34.35 & 59.69 & 96.64\scriptsize\textit{ms/token} \\
    10 & Yi-34B-Chat & 72.75 & 68.76 & 50.88 & 70.01 & 29.81 & 58.44 & 67.96\scriptsize\textit{ms/token} \\
    \midrule
    \multicolumn{9}{c}{\emph{Ensemble Results for the Above Two Models (Mixtral and Yi)}} \\
    \midrule
    11 & {LLM Blender (PairRanker)} & 72.69 & 69.59 & 51.70 & 72.37 & 32.24 & 59.72 & 105.21\scriptsize\textit{ms/token} \\
    12 & {OAssistRM} & 73.34 & 70.15 & 51.91 & 72.79 & 30.69 & 59.78 & 99.75\scriptsize\textit{ms/token} \\
    13 & {UltraRM} & 69.49 & 71.09 & 52.27 & 73.82 & 32.36 & 59.81 & 114.57\scriptsize\textit{ms/token} \\
    14 & \textsc{GaC} (\textit{\textbf{ours}}) & \tightcellcolor{beaublue}{74.83} & \tightcellcolor{beaublue}{71.21} & \tightcellcolor{beaublue}{52.64} & \tightcellcolor{beaublue}{75.60} & \tightcellcolor{beaublue}{33.52} & \tightcellcolor{beaublue}{61.56} & 98.13\scriptsize\textit{ms/token} \\
    \bottomrule
\end{tabular}
\caption{Results of comparison with other methods. Upper and lower halves represent different ensemble combinations. Blue indicates the best result for each ensemble. For id 8, \textsc{GaC} bottom right shows ensembled token proportion, top right shows threshold. For id 6 and 14, we ensemble at every generation step.}
\label{table3}
\end{table*}

\textbf{Baselines.} We compared \textsc{GaC} with existing methods. First, we considered LLM Blender \citep{jiang2023llm}, which employs PairRanker to rank the outputs of candidate LLMs and GenFuser to fuse these outputs. However, we found that GenFuser refused to answer a significant proportion of the questions in our chosen benchmarks. We therefore only used PairRanker to ensure fairness. We also included other rankers, such as OAssistRM\footnote{\footnotesize \href{https://huggingface.co/OpenAssistant/reward-model-deberta-v3-large-v2}{https://huggingface.co/OpenAssistant/reward-model}} \citep{kopf2024openassistant} and UltraRM \citep{cui2023ultrafeedback}, which we ran in parallel on the GPUs hosting the ensemble LLMs to ensure low latency. These rankers scored the outputs and selected the best answers. Furthermore, we included the FuseLLM \citep{wan2024knowledge} (OpenChat-3.5-7B-Solar\footnote{\footnotesize \href{https://huggingface.co/FuseAI/OpenChat-3.5-7B-Solar}{https://huggingface.co/FuseAI/OpenChat-3.5-7B-Solar}}), which uses probability information from multiple models during training for distillation.

\vspace{0.5\baselineskip}
\noindent\textbf{Models for Ensemble.} We chose two sets of models of different sizes. The smaller models included openchat-3.5 \citep{wang2023openchat} and Nous-Hermes-2-SOLAR-10.7B \citep{nousSOLAR} (teacher models for OpenChat-3.5-7B-Solar distillation). Larger models included Mixtral-8x7B-Instruct-v0.1 and Yi-34B-Chat \citep{jiang2024mixtral,young2024yi}.

\vspace{0.5\baselineskip}
\noindent\textbf{Experimental Results.} We ensemble the above two sets of LLMs using both our and baseline methods, and present the results in Tab.\ref{table3}. Our method showed superior performance with the lowest latency for both combinations (row ids 6 and 14). For row id 8, we used openchat-3.5 as the gate model with a threshold of 0.5 (Eq.\ref{eq5}), resulting in only $7.68\%$ of tokens being ensembled. This slightly increased the latency from 28.01 to 31.34 ms/token, but achieved a performance close to that of SOLAR-10.7B (average score of 58.01 vs. 58.24), whose latency is 50.89 ms/token, further demonstrating the effectiveness of our method.

\subsection{Breaking the Ceiling}
\label{sec4-5}
\begin{table*}[!ht]
\centering
\footnotesize
\setlength{\tabcolsep}{2.5pt} 
\begin{tabular}{>{\raggedright\arraybackslash}p{0.4cm}>{\raggedright\arraybackslash}p{3.5cm}cccccccc}
    \toprule
    \textbf{Id} & \textbf{Models} & \textbf{MMLU} & \textbf{GSM8K} & \textbf{BBH} & \textbf{TriviaQA} & \textbf{NQ} & \textbf{Avg.} & \textbf{Date} & \textbf{Latency} \\
    \midrule
    1 & Yi-34B-Chat & 72.75 & 68.76 & 50.88 & 70.01 & 29.81 & 58.44 & 2023/11/08 & 67.96\scriptsize\textit{ms/token} \\
    2 & Mixtral-8x7B-Instruct-v0.1 & 70.89 & 66.82 & 49.84 & 76.54 & 34.35 & 59.69 & 2023/12/11 & 96.64\scriptsize\textit{ms/token} \\
    3 & Qwen1.5-72B-Chat & 77.79 & 83.33 & 48.94 & 65.69 & 27.02 & 60.55 & 2024/02/04 & 102.11\scriptsize\textit{ms/token} \\
    4 & Llama-3-70B-Instruct & 79.68 & 90.00 & 57.13 & 79.12 & 35.57 & 68.30 & 2024/04/18 & 150.32\scriptsize\textit{ms/token} \\
    5 & Qwen2-72B-Instruct & 82.30 & 89.70 & 62.57 & 73.58 & 33.11 & 68.25 & 2024/06/07 & 113.91\scriptsize\textit{ms/token} \\
    \midrule
    \multicolumn{10}{c}{\emph{Ensemble the Above Models with }\textsc{GaC}} \\
    \midrule
    6 & Yi + Mixtral & 74.83 & 71.21 & 52.64 & 75.60 & 33.52 & 61.56\textbf{$\uparrow$3.13\%} & \textasciitilde2023/12/11 & 98.13\scriptsize\textit{ms/token} \\
    7 & Qwen1.5-72B + Yi & 79.83 & 77.27 & 52.05 & 70.88 & 33.80 & 62.77\textbf{$\uparrow$3.65\%} & \textasciitilde2024/02/04 & 103.69\scriptsize\textit{ms/token} \\
    8 & Qwen1.5-72B + Mixtral & 79.55 & 75.76 & 54.19 & 75.71 & 31.09 & 63.26\textbf{$\uparrow$4.47\%} & \textasciitilde2024/02/04 & 112.83\scriptsize\textit{ms/token} \\
    9 & Llama-3 + Qwen1.5-72B & 81.49 & 87.06 & 56.73 & 78.60 & 36.01 & 67.98\textbf{$\downarrow$0.47\%} & \textasciitilde2024/04/18 & 153.96\scriptsize\textit{ms/token} \\
    10 & Qwen2-72B + Llama-3 & 83.54 & 90.91 & 63.99 & 79.29 & 37.65 & 71.08\textbf{$\uparrow$4.06\%} & \textasciitilde2024/06/07 & 151.56\scriptsize\textit{ms/token} \\
    \midrule
    \multicolumn{10}{c}{\emph{Task-specific Top-2 Model Ensemble with }\textsc{GaC}} \\
    \midrule
    11 & Top-2 (\textasciitilde2024/04/18)& 81.49 & 87.96 & 58.64 & 80.84 & 37.95 & 69.38\textbf{$\uparrow$1.58\%} & \textasciitilde2024/04/18 & - \\
    12 & Top-2 (\textasciitilde2024/06/07)& 83.54 & 90.91 & 63.99 & 80.84 & 37.95 & 71.45\textbf{$\uparrow$4.61\%} & \textasciitilde2024/06/07 & - \\
    \bottomrule
\end{tabular}
\caption{Ensemble of available SOTA LLMs from different periods. The top part lists the individual models, while the bottom part shows the ensemble results (model names abbreviated). \textbf{$\uparrow$} indicates the percentage improvement over the individual models.}
\label{table4}
\end{table*}

\begin{figure}
  \centering
  \includegraphics[width=\columnwidth]{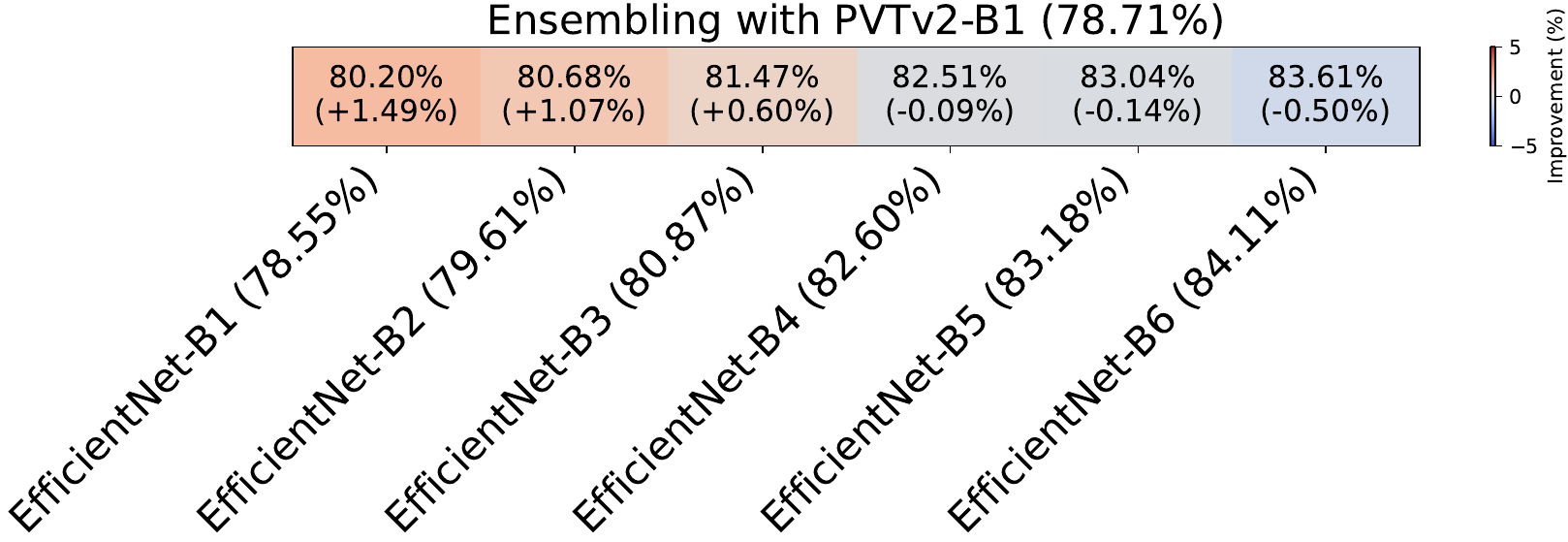}
  \caption{Ensemble results of CV models with different accuracy gaps on ImageNet. Models' accuracies are next to their names. Each cell shows the ensemble accuracy, with the improvement over the best single model in parentheses.}
  \label{fig_5}
\end{figure}

In this experiment, we aimed to break the performance ceiling of the open source LLM community at different times. We chose the SOTA LLMs released between November 2023 and June 2024, as listed in the upper part of Tab.\ref{table4}, excluding models with more than 100 billion parameters due to hardware limitations. We then ensemble the SOTA LLMs available at different times, as shown in row ids 6-10, and observe an improvement of \textbf{3.13\%} to \textbf{4.47\%} over the best single model at each time. An exception is 2024/04/18, when Llama-3-70B-Instruct was released and significantly improved performance over the previous SOTA LLMs (average score increased from 60.55 of Qwen1.5-72B-Chat to 68.30), resulting in a drop in performance after ensemble due to the large gap.

However, with the release of Qwen2-72B-Instruct, which showed comparable performance to Llama-3-70B-Instruct, the ensemble again led to significant improvements (row id 10). In rows 11 and 12, we ensemble the top two best-performing models for each benchmark at the two most recent times, including the challenging time of 2024/04/18, and observe performance gains with this task-specific top-two ensemble even on 04/18 (row id 11). Finally, row id 12 shows the \textbf{best} results available for the open source community on 2024/06/07. By pushing the boundaries of the community, we can narrow the gap with proprietary models and promote the democratization of LLMs.\footnote{We also tested translation and summarization in Appx.\ref{sec:appendix-TS}.}

\begin{table*}[t]
\centering
\footnotesize
\setlength{\tabcolsep}{3pt} 
\begin{tabular}{>{\raggedright\arraybackslash}p{0.2cm}>{\raggedright\arraybackslash}p{4.55cm}cccccccc}
    \toprule
    \textbf{Id} & \textbf{Models} & \textbf{Threshold} & \textbf{MMLU} & \textbf{GSM8K} & \textbf{BBH} & \textbf{TriviaQA} & \textbf{NQ} & \textbf{Avg.} & \textbf{Latency} \\
    \midrule
    1 & Llama-3-70B-Instruct & - & 79.68 & 90.00 & 57.13 & 79.12 & 35.57 & 68.30 & 150.32\scriptsize\textit{ms/token} \\
    2 & Llama-3-8B-Instruct & - & 65.08 & 76.26 & 44.72 & 67.67 & 26.48 & 56.04 & 34.30\scriptsize\textit{ms/token} \\
    3 & Phi-3-mini-4k-instruct & - & 67.11 & 79.00 & 47.26 & 55.78 & 17.98 & 53.43 & 31.82\scriptsize\textit{ms/token} \\
    4 & Qwen1.5-72B-chat & - & 77.79 & 83.33 & 48.94 & 65.69 & 27.02 & 60.55 & 102.11\scriptsize\textit{ms/token} \\
    5 & Qwen1.5-32B-Chat & - & 75.12 & 75.97 & 53.89 & 62.57 & 22.96 & 58.10 & 59.01\scriptsize\textit{ms/token} \\
    \midrule
    \multicolumn{10}{c}{\textit{Ensemble with threshold to match Qwen1.5-\textbf{72B}-Chat performance (avg. 60.55)}} \\
    \midrule
    6 & Llama-3-8B + Llama-3-70B$_{9.30\%}$ & 0.5 & 69.51 & 82.86 & 47.65 & 74.35 & 33.74 & 61.62 & 68.97\scriptsize\textit{ms/token} \\
    7 & Qwen1.5-32B + Qwen1.5-72B$_{6.31\%}$ & 0.5 & 75.53 & 81.82 & 55.87 & 63.91 & 27.69 & 60.96 & 77.86\scriptsize\textit{ms/token} \\
    \midrule
    \multicolumn{10}{c}{\textit{Ensemble with threshold to match Qwen1.5-\textbf{32B}-Chat performance (avg. 58.10)}} \\
    \midrule
    8 & Llama-3-8B + Llama-3-70B$_{6.98\%}$ & 0.45 & 68.06 & 81.79 & 46.66 & 73.43 & 33.74 & 60.74 & 58.99\scriptsize\textit{ms/token} \\
    9 & Phi-3 + Llama-3-70B$_{7.59\%}$ & 0.5 & 68.46 & 78.57 & 50.34 & 69.08 & 29.51 & 59.19 & 51.61\scriptsize\textit{ms/token} \\
    \bottomrule
\end{tabular}
\caption{Thresholded ensemble results. The top lists individual models, while the bottom shows ensemble combinations (model names abbreviated). The percentage in the bottom right of the combination names represents the proportion of tokens ensembled.}
\label{table5}
\end{table*}

\begin{table}[t]
\centering
\footnotesize
\setlength{\tabcolsep}{0.5pt} 
\begin{tabular}{>{\raggedright\arraybackslash}p{2.8cm}ccc}
    \toprule
    \textbf{Models} & \textbf{Threshold} & \textbf{MT-Bench} & \textbf{Latency} \\
    \midrule
    Llama-3-8b-Instruct & - & 8.03 & 33.75\scriptsize\textit{ms/token} \\
    Llama-3-70b-Instruct & - & 8.80 & 128.50\scriptsize\textit{ms/token} \\
    \midrule
    Qwen1.5-72b-chat & - & 8.33 & 87.39\scriptsize\textit{ms/token} \\
    \textsc{GaC} (Llama-3-8b + Llama-3-70b)$_{10.10\%}$ & 0.55 & 8.34 & 69.28\scriptsize\textit{ms/token} \\
    \midrule
    Qwen1.5-32b-chat & - & 8.12 & 58.49\scriptsize\textit{ms/token} \\
    \textsc{GaC} (Llama-3-8b + Llama-3-70b)$_{7.24\%}$ & 0.5 & 8.16 & 57.28\scriptsize\textit{ms/token} \\
    \bottomrule
\end{tabular}
\caption{Thresholded ensemble on MT-Bench. \textsc{GaC} shows the ensemble combinations (model names abbreviated), with the proportion of tokens ensembled shown at the bottom right.}
\label{table6}
\end{table}

Since ensembling models with large performance differences could lead to performance degradation (row id 9 in Tab.\ref{table4}), we also tested this hypothesis with CV models. We ensemble PVTv2-B1 \cite{wang2021pvtv2} with different sizes of EfficientNet \citep{tan2019efficientnet} on ImageNet \citep{deng2009imagenet} by averaging their outputs, as shown in Fig.\ref{fig_5}. We observed that as the accuracy gap between the two models increased, the ensemble gains decreased and eventually became negative. This suggests that it is advisable to ensemble models with similar levels of performance.

\subsection{Ensemble with Threshold}
\label{sec4-6}

In this experiment, we used the thresholded ensemble (Sec.\ref{sec3-4}) to explore variations in latency and performance. We selected models of different sizes \citep{abdin2024phi,meta2024llama3,bai2023qwen}, listed in the upper part of Tab.\ref{table5}, pairing a smaller model with a larger model and using the smaller model as the gate model for the ensemble. We aimed to match the performance (average score) of Qwen1.5-72B-chat and Qwen1.5-32B-chat with our ensemble, but with lower latency, and the results are shown in row ids 6-9. Interestingly, even when combining the two Qwen models themselves with a threshold of 0.5 (row id 7), where 6.31\% of the tokens were ensembled, we observed slightly higher performance than Qwen1.5-72B-chat (average score increased from 60.55 to 60.96) and lower latency (102.11 to 77.86 ms/token). We believe this is a promising new way to speed up inference.

We also observed lower latency in row ids 6, 8 and 9 of Tab.\ref{table5} with comparable performance to Qwen1.5-72B-chat or Qwen1.5-32B-chat. In addition, similar trends were observed in the MT-Bench \citep{zheng2023judging} using LLM as judge (GPT-4-0613) in Tab.\ref{table6}, with scores calculated using FastChat\footnote{\footnotesize \href{https://github.com/lm-sys/FastChat}{https://github.com/lm-sys/FastChat}} in our local environment. For the combinations of Llama-3-8B-Instruct or Phi-3-mini-4k-instruct with Llama-3-70B-Instruct, we directly adopted the output of the larger model if the probability of the smaller model was below the threshold. This is based on our observations in Fig.\ref{fig_5}, where ensembling models with large performance gaps resulted in reduced performance.

\section{Conclusion}
In this paper, we present a token-level ensembling framework called \textsc{GaC}, which fully exploits the probability information at each generation step. In our experiments, we have surpassed the performance ceiling of open-source SOTA LLMs available at different time periods (Sec.\ref{sec4-5}), further narrowing the gap between open-source and proprietary models. This  progress promotes the democratization of LLMs and provides new motivations for future research, enabling better exploitation of collective intelligence. In addition, we experimented with ensembling just a few tokens and found that this approach can achieve better performance with lower latency (Sec.\ref{sec4-6}), opening up new avenues for accelerating inference.

\section*{Contemporaneous Works}
We have noticed several contemporaneous works related to our research, all of which aim to address the vocabulary discrepancy between different models. \citet{xu2024bridging} proposed EVA, which trains a projection matrix between each pair of LLMs, using the overlapping tokens from their vocabularies as a bridge. DEEPEN \citep{huang2024enabling} converts the output probabilities to a relative representation using anchor tokens before ensembling, and then inverts back to the original model's vocabulary space using gradient descent, which requires an additional $7\%$ to $29\%$ of time per generation step. In contrast, our method requires no additional training and only a single matrix multiplication and tokenization for each model during ensembling, with minimal time cost.

\section*{Limitations}
Like other ensemble methods, the approach proposed in this paper requires more computational resources. Although different models can be run in parallel on separate GPUs, so that latency only depends on the slowest model, the overall computational load is additive, raising the threshold for use.
\bibliography{custom}

\begin{thebibliography}{57}
\providecommand{\natexlab}[1]{#1}

\bibitem[{Abdin et~al.(2024)Abdin, Jacobs, Awan, Aneja, Awadallah, Awadalla, Bach, Bahree, Bakhtiari, Behl et~al.}]{abdin2024phi}
Marah Abdin, Sam~Ade Jacobs, Ammar~Ahmad Awan, Jyoti Aneja, Ahmed Awadallah, Hany Awadalla, Nguyen Bach, Amit Bahree, Arash Bakhtiari, Harkirat Behl, et~al. 2024.
\newblock Phi-3 technical report: A highly capable language model locally on your phone.
\newblock \emph{arXiv preprint arXiv:2404.14219}.

\bibitem[{Achiam et~al.(2023)Achiam, Adler, Agarwal, Ahmad, Akkaya, Aleman, Almeida, Altenschmidt, Altman, Anadkat et~al.}]{achiam2023gpt}
Josh Achiam, Steven Adler, Sandhini Agarwal, Lama Ahmad, Ilge Akkaya, Florencia~Leoni Aleman, Diogo Almeida, Janko Altenschmidt, Sam Altman, Shyamal Anadkat, et~al. 2023.
\newblock Gpt-4 technical report.
\newblock \emph{arXiv preprint arXiv:2303.08774}.

\bibitem[{Alibaba(2024)}]{qwen2blog}
Alibaba. 2024.
\newblock \href {https://qwenlm.github.io/blog/qwen2/} {Qwen-2: Blog post}.
\newblock Accessed: 2024-06-07.

\bibitem[{AllenAI(2024)}]{allenai2023olmo}
AllenAI. 2024.
\newblock \href {https://blog.allenai.org/hello-olmo-a-truly-open-llm-43f7e7359222} {Hello olmo: A truly open llm}.
\newblock Accessed: 2024-06-07.

\bibitem[{Almazrouei et~al.(2023)Almazrouei, Alobeidli, Alshamsi, Cappelli, Cojocaru, Debbah, Goffinet, Hesslow, Launay, Malartic et~al.}]{almazrouei2023falcon}
Ebtesam Almazrouei, Hamza Alobeidli, Abdulaziz Alshamsi, Alessandro Cappelli, Ruxandra Cojocaru, M{\'e}rouane Debbah, {\'E}tienne Goffinet, Daniel Hesslow, Julien Launay, Quentin Malartic, et~al. 2023.
\newblock The falcon series of open language models.
\newblock \emph{arXiv preprint arXiv:2311.16867}.

\bibitem[{Bai et~al.(2023)Bai, Bai, Chu, Cui, Dang, Deng, Fan, Ge, Han, Huang et~al.}]{bai2023qwen}
Jinze Bai, Shuai Bai, Yunfei Chu, Zeyu Cui, Kai Dang, Xiaodong Deng, Yang Fan, Wenbin Ge, Yu~Han, Fei Huang, et~al. 2023.
\newblock Qwen technical report.
\newblock \emph{arXiv preprint arXiv:2309.16609}.

\bibitem[{Barad et~al.(2023)Barad, Aidova, and Gorbachev}]{barad2023leveraging}
Haim Barad, Ekaterina Aidova, and Yury Gorbachev. 2023.
\newblock Leveraging speculative sampling and kv-cache optimizations together for generative ai using openvino.
\newblock \emph{arXiv preprint arXiv:2311.04951}.

\bibitem[{Bojar et~al.(2016)Bojar, Chatterjee, Federmann, Graham, Haddow, Huck, Jimeno~Yepes, Koehn, Logacheva, Monz, Negri, Neveol, Neves, Popel, Post, Rubino, Scarton, Specia, Turchi, Verspoor, and Zampieri}]{bojar-EtAl:2016:WMT1}
Ond~{r}ej Bojar, Rajen Chatterjee, Christian Federmann, Yvette Graham, Barry Haddow, Matthias Huck, Antonio Jimeno~Yepes, Philipp Koehn, Varvara Logacheva, Christof Monz, Matteo Negri, Aurelie Neveol, Mariana Neves, Martin Popel, Matt Post, Raphael Rubino, Carolina Scarton, Lucia Specia, Marco Turchi, Karin Verspoor, and Marcos Zampieri. 2016.
\newblock \href {http://www.aclweb.org/anthology/W/W16/W16-2301} {Findings of the 2016 conference on machine translation}.
\newblock In \emph{Proceedings of the First Conference on Machine Translation}, pages 131--198, Berlin, Germany. Association for Computational Linguistics.

\bibitem[{Bojar et~al.(2014)Bojar, Buck, Federmann, Haddow, Koehn, Leveling, Monz, Pecina, Post, Saint-Amand, Soricut, Specia, and Tamchyna}]{bojar-EtAl:2014:W14-33}
Ondrej Bojar, Christian Buck, Christian Federmann, Barry Haddow, Philipp Koehn, Johannes Leveling, Christof Monz, Pavel Pecina, Matt Post, Herve Saint-Amand, Radu Soricut, Lucia Specia, and Ale~{s} Tamchyna. 2014.
\newblock \href {http://www.aclweb.org/anthology/W/W14/W14-3302} {Findings of the 2014 workshop on statistical machine translation}.
\newblock In \emph{Proceedings of the Ninth Workshop on Statistical Machine Translation}, pages 12--58, Baltimore, Maryland, USA. Association for Computational Linguistics.

\bibitem[{Chen et~al.(2022)Chen, Guo, Zeng, Xiong, and Dong}]{chen2022repghost}
Chengpeng Chen, Zichao Guo, Haien Zeng, Pengfei Xiong, and Jian Dong. 2022.
\newblock Repghost: a hardware-efficient ghost module via re-parameterization.
\newblock \emph{arXiv preprint arXiv:2211.06088}.

\bibitem[{Cobbe et~al.(2021)Cobbe, Kosaraju, Bavarian, Chen, Jun, Kaiser, Plappert, Tworek, Hilton, Nakano et~al.}]{cobbe2021training}
Karl Cobbe, Vineet Kosaraju, Mohammad Bavarian, Mark Chen, Heewoo Jun, Lukasz Kaiser, Matthias Plappert, Jerry Tworek, Jacob Hilton, Reiichiro Nakano, et~al. 2021.
\newblock Training verifiers to solve math word problems.
\newblock \emph{arXiv preprint arXiv:2110.14168}.

\bibitem[{Cui et~al.(2023)Cui, Yuan, Ding, Yao, Zhu, Ni, Xie, Liu, and Sun}]{cui2023ultrafeedback}
Ganqu Cui, Lifan Yuan, Ning Ding, Guanming Yao, Wei Zhu, Yuan Ni, Guotong Xie, Zhiyuan Liu, and Maosong Sun. 2023.
\newblock Ultrafeedback: Boosting language models with high-quality feedback.
\newblock \emph{arXiv preprint arXiv:2310.01377}.

\bibitem[{Databricks(2024)}]{databricks2023dbrx}
Databricks. 2024.
\newblock \href {https://www.databricks.com/blog/introducing-dbrx-new-state-art-open-llm} {Introducing dbrx: A new state-of-the-art open llm}.
\newblock Accessed: 2024-06-07.

\bibitem[{Deng et~al.(2009)Deng, Dong, Socher, Li, Li, and Fei-Fei}]{deng2009imagenet}
Jia Deng, Wei Dong, Richard Socher, Li-Jia Li, Kai Li, and Li~Fei-Fei. 2009.
\newblock Imagenet: A large-scale hierarchical image database.
\newblock In \emph{2009 IEEE conference on computer vision and pattern recognition}, pages 248--255. Ieee.

\bibitem[{Enomoro and Eda(2021)}]{enomoro2021learning}
Shohei Enomoro and Takeharu Eda. 2021.
\newblock Learning to cascade: Confidence calibration for improving the accuracy and computational cost of cascade inference systems.
\newblock In \emph{Proceedings of the AAAI Conference on Artificial Intelligence}, 8, pages 7331--7339.

\bibitem[{Gao et~al.(2023)Gao, Tow, Abbasi, Biderman, Black, DiPofi, Foster, Golding, Hsu, Le~Noac'h, Li, McDonell, Muennighoff, Ociepa, Phang, Reynolds, Schoelkopf, Skowron, Sutawika, Tang, Thite, Wang, Wang, and Zou}]{eval-harness}
Leo Gao, Jonathan Tow, Baber Abbasi, Stella Biderman, Sid Black, Anthony DiPofi, Charles Foster, Laurence Golding, Jeffrey Hsu, Alain Le~Noac'h, Haonan Li, Kyle McDonell, Niklas Muennighoff, Chris Ociepa, Jason Phang, Laria Reynolds, Hailey Schoelkopf, Aviya Skowron, Lintang Sutawika, Eric Tang, Anish Thite, Ben Wang, Kevin Wang, and Andy Zou. 2023.
\newblock \href {https://doi.org/10.5281/zenodo.10256836} {A framework for few-shot language model evaluation}.

\bibitem[{Guo et~al.(2017)Guo, Pleiss, Sun, and Weinberger}]{guo2017calibration}
Chuan Guo, Geoff Pleiss, Yu~Sun, and Kilian~Q Weinberger. 2017.
\newblock On calibration of modern neural networks.
\newblock In \emph{International conference on machine learning}, pages 1321--1330. PMLR.

\bibitem[{Hendrycks et~al.(2020)Hendrycks, Burns, Basart, Zou, Mazeika, Song, and Steinhardt}]{hendrycks2020measuring}
Dan Hendrycks, Collin Burns, Steven Basart, Andy Zou, Mantas Mazeika, Dawn Song, and Jacob Steinhardt. 2020.
\newblock Measuring massive multitask language understanding.
\newblock \emph{arXiv preprint arXiv:2009.03300}.

\bibitem[{Huang et~al.(2016)Huang, Li, Pleiss, Liu, Hopcroft, and Weinberger}]{huang2016snapshot}
Gao Huang, Yixuan Li, Geoff Pleiss, Zhuang Liu, John~E Hopcroft, and Kilian~Q Weinberger. 2016.
\newblock Snapshot ensembles: Train 1, get m for free.
\newblock In \emph{International Conference on Learning Representations}.

\bibitem[{Huang et~al.(2024)Huang, Feng, Li, Xiang, Wang, Qin, and Liu}]{huang2024enabling}
Yichong Huang, Xiaocheng Feng, Baohang Li, Yang Xiang, Hui Wang, Bing Qin, and Ting Liu. 2024.
\newblock Enabling ensemble learning for heterogeneous large language models with deep parallel collaboration.
\newblock \emph{arXiv preprint arXiv:2404.12715}.

\bibitem[{HuggingFace(2024)}]{huggingface2024}
HuggingFace. 2024.
\newblock Transformers documentation: Model parallelism.
\newblock \url{https://huggingface.co/docs/transformers/v4.15.0/parallelism}.
\newblock Accessed: 2024-06-07.

\bibitem[{Jazbec et~al.(2024)Jazbec, Allingham, Zhang, and Nalisnick}]{jazbec2024towards}
Metod Jazbec, James Allingham, Dan Zhang, and Eric Nalisnick. 2024.
\newblock Towards anytime classification in early-exit architectures by enforcing conditional monotonicity.
\newblock \emph{Advances in Neural Information Processing Systems}, 36.

\bibitem[{Jiang et~al.(2024)Jiang, Sablayrolles, Roux, Mensch, Savary, Bamford, Chaplot, Casas, Hanna, Bressand et~al.}]{jiang2024mixtral}
Albert~Q Jiang, Alexandre Sablayrolles, Antoine Roux, Arthur Mensch, Blanche Savary, Chris Bamford, Devendra~Singh Chaplot, Diego de~las Casas, Emma~Bou Hanna, Florian Bressand, et~al. 2024.
\newblock Mixtral of experts.
\newblock \emph{arXiv preprint arXiv:2401.04088}.

\bibitem[{Jiang et~al.(2023)Jiang, Ren, and Lin}]{jiang2023llm}
Dongfu Jiang, Xiang Ren, and Bill~Yuchen Lin. 2023.
\newblock Llm-blender: Ensembling large language models with pairwise ranking and generative fusion.
\newblock In \emph{Proceedings of the 61st Annual Meeting of the Association for Computational Linguistics (Volume 1: Long Papers)}, pages 14165--14178.

\bibitem[{Joshi et~al.(2017)Joshi, Choi, Weld, and Zettlemoyer}]{joshi2017triviaqa}
Mandar Joshi, Eunsol Choi, Daniel~S Weld, and Luke Zettlemoyer. 2017.
\newblock Triviaqa: A large scale distantly supervised challenge dataset for reading comprehension.
\newblock In \emph{Proceedings of the 55th Annual Meeting of the Association for Computational Linguistics (Volume 1: Long Papers)}, pages 1601--1611.

\bibitem[{Kim et~al.(2023)Kim, Park, Kim, Lee, Song, Kim, Kim, Kim, Lee, Kim et~al.}]{kim2023solar}
Dahyun Kim, Chanjun Park, Sanghoon Kim, Wonsung Lee, Wonho Song, Yunsu Kim, Hyeonwoo Kim, Yungi Kim, Hyeonju Lee, Jihoo Kim, et~al. 2023.
\newblock Solar 10.7 b: Scaling large language models with simple yet effective depth up-scaling.
\newblock \emph{arXiv preprint arXiv:2312.15166}.

\bibitem[{K{\"o}pf et~al.(2024)K{\"o}pf, Kilcher, von R{\"u}tte, Anagnostidis, Tam, Stevens, Barhoum, Nguyen, Stanley, Nagyfi et~al.}]{kopf2024openassistant}
Andreas K{\"o}pf, Yannic Kilcher, Dimitri von R{\"u}tte, Sotiris Anagnostidis, Zhi~Rui Tam, Keith Stevens, Abdullah Barhoum, Duc Nguyen, Oliver Stanley, Rich{\'a}rd Nagyfi, et~al. 2024.
\newblock Openassistant conversations-democratizing large language model alignment.
\newblock \emph{Advances in Neural Information Processing Systems}, 36.

\bibitem[{Krizhevsky et~al.(2017)Krizhevsky, Sutskever, and Hinton}]{krizhevsky2017imagenet}
Alex Krizhevsky, Ilya Sutskever, and Geoffrey~E Hinton. 2017.
\newblock Imagenet classification with deep convolutional neural networks.
\newblock \emph{Communications of the ACM}, 60(6):84--90.

\bibitem[{Kwiatkowski et~al.(2019)Kwiatkowski, Palomaki, Redfield, Collins, Parikh, Alberti, Epstein, Polosukhin, Devlin, Lee et~al.}]{kwiatkowski2019natural}
Tom Kwiatkowski, Jennimaria Palomaki, Olivia Redfield, Michael Collins, Ankur Parikh, Chris Alberti, Danielle Epstein, Illia Polosukhin, Jacob Devlin, Kenton Lee, et~al. 2019.
\newblock Natural questions: a benchmark for question answering research.
\newblock \emph{Transactions of the Association for Computational Linguistics}, 7:453--466.

\bibitem[{Kwon et~al.(2023)Kwon, Li, Zhuang, Sheng, Zheng, Yu, Gonzalez, Zhang, and Stoica}]{kwon2023efficient}
Woosuk Kwon, Zhuohan Li, Siyuan Zhuang, Ying Sheng, Lianmin Zheng, Cody~Hao Yu, Joseph Gonzalez, Hao Zhang, and Ion Stoica. 2023.
\newblock Efficient memory management for large language model serving with pagedattention.
\newblock In \emph{Proceedings of the 29th Symposium on Operating Systems Principles}, pages 611--626.

\bibitem[{Lu et~al.(2023)Lu, Yuan, Lin, Lin, Yuan, Zhou, and Zhou}]{lu2023routing}
Keming Lu, Hongyi Yuan, Runji Lin, Junyang Lin, Zheng Yuan, Chang Zhou, and Jingren Zhou. 2023.
\newblock Routing to the expert: Efficient reward-guided ensemble of large language models.
\newblock \emph{arXiv preprint arXiv:2311.08692}.

\bibitem[{Mehta et~al.(2024)Mehta, Sekhavat, Cao, Horton, Jin, Sun, Mirzadeh, Najibi, Belenko, Zatloukal et~al.}]{mehta2024openelm}
Sachin Mehta, Mohammad~Hossein Sekhavat, Qingqing Cao, Maxwell Horton, Yanzi Jin, Chenfan Sun, Iman Mirzadeh, Mahyar Najibi, Dmitry Belenko, Peter Zatloukal, et~al. 2024.
\newblock Openelm: An efficient language model family with open-source training and inference framework.
\newblock \emph{arXiv preprint arXiv:2404.14619}.

\bibitem[{Meta(2024)}]{meta2024llama3}
Meta. 2024.
\newblock \href {https://ai.meta.com/blog/meta-llama-3/} {Meta ai blog: Meta llama 3}.
\newblock Accessed: 2024-06-07.

\bibitem[{Moritz et~al.(2018)Moritz, Nishihara, Wang, Tumanov, Liaw, Liang, Elibol, Yang, Paul, Jordan et~al.}]{moritz2018ray}
Philipp Moritz, Robert Nishihara, Stephanie Wang, Alexey Tumanov, Richard Liaw, Eric Liang, Melih Elibol, Zongheng Yang, William Paul, Michael~I Jordan, et~al. 2018.
\newblock Ray: A distributed framework for emerging $\{$AI$\}$ applications.
\newblock In \emph{13th USENIX symposium on operating systems design and implementation (OSDI 18)}, pages 561--577.

\bibitem[{Narayan et~al.(2018)Narayan, Cohen, and Lapata}]{narayan2018don}
Shashi Narayan, Shay~B Cohen, and Mirella Lapata. 2018.
\newblock Don't give me the details, just the summary! topic-aware convolutional neural networks for extreme summarization.
\newblock \emph{arXiv preprint arXiv:1808.08745}.

\bibitem[{NousResearch(2024{\natexlab{a}})}]{nousmixtral}
NousResearch. 2024{\natexlab{a}}.
\newblock Nous-hermes-2-mixtral-8x7b-dpo.
\newblock \url{https://huggingface.co/NousResearch/Nous-Hermes-2-Mixtral-8x7B-DPO}.
\newblock Accessed: 2024-06-07.

\bibitem[{NousResearch(2024{\natexlab{b}})}]{nousSOLAR}
NousResearch. 2024{\natexlab{b}}.
\newblock Nous-hermes-2-solar-10.7b.
\newblock \url{https://huggingface.co/NousResearch/Nous-Hermes-2-SOLAR-10.7B}.
\newblock Accessed: 2024-06-07.

\bibitem[{Oxford(2018)}]{oxford5000}
Oxford. 2018.
\newblock \href {https://www.oxfordlearnersdictionaries.com/wordlists/oxford3000-5000} {The oxford 5000}.
\newblock Accessed: 2024-06-07.

\bibitem[{Rasley et~al.(2020)Rasley, Rajbhandari, Ruwase, and He}]{rasley2020deepspeed}
Jeff Rasley, Samyam Rajbhandari, Olatunji Ruwase, and Yuxiong He. 2020.
\newblock Deepspeed: System optimizations enable training deep learning models with over 100 billion parameters.
\newblock In \emph{Proceedings of the 26th ACM SIGKDD International Conference on Knowledge Discovery \& Data Mining}, pages 3505--3506.

\bibitem[{Schoenegger et~al.(2024)Schoenegger, Tuminauskaite, Park, and Tetlock}]{schoenegger2024wisdom}
Philipp Schoenegger, Indre Tuminauskaite, Peter~S Park, and Philip~E Tetlock. 2024.
\newblock Wisdom of the silicon crowd: Llm ensemble prediction capabilities match human crowd accuracy.
\newblock \emph{arXiv preprint arXiv:2402.19379}.

\bibitem[{Sennrich et~al.(2015)Sennrich, Haddow, and Birch}]{sennrich2015neural}
Rico Sennrich, Barry Haddow, and Alexandra Birch. 2015.
\newblock Neural machine translation of rare words with subword units.
\newblock \emph{arXiv preprint arXiv:1508.07909}.

\bibitem[{Shnitzer et~al.(2023)Shnitzer, Ou, Silva, Soule, Sun, Solomon, Thompson, and Yurochkin}]{shnitzer2023large}
Tal Shnitzer, Anthony Ou, Mirian Silva, Kate Soule, Yuekai Sun, Justin Solomon, Neil Thompson, and Mikhail Yurochkin. 2023.
\newblock Large language model routing with benchmark datasets.
\newblock In \emph{Annual Conference on Neural Information Processing Systems}.

\bibitem[{Suzgun et~al.(2023)Suzgun, Scales, Sch{\"a}rli, Gehrmann, Tay, Chung, Chowdhery, Le, Chi, Zhou et~al.}]{suzgun2023challenging}
Mirac Suzgun, Nathan Scales, Nathanael Sch{\"a}rli, Sebastian Gehrmann, Yi~Tay, Hyung~Won Chung, Aakanksha Chowdhery, Quoc Le, Ed~Chi, Denny Zhou, et~al. 2023.
\newblock Challenging big-bench tasks and whether chain-of-thought can solve them.
\newblock In \emph{Findings of the Association for Computational Linguistics: ACL 2023}, pages 13003--13051.

\bibitem[{Tan and Le(2019)}]{tan2019efficientnet}
Mingxing Tan and Quoc Le. 2019.
\newblock Efficientnet: Rethinking model scaling for convolutional neural networks.
\newblock In \emph{International conference on machine learning}, pages 6105--6114. PMLR.

\bibitem[{Touvron et~al.(2023)Touvron, Martin, Stone, Albert, Almahairi, Babaei, Bashlykov, Batra, Bhargava, Bhosale et~al.}]{touvron2023llama}
Hugo Touvron, Louis Martin, Kevin Stone, Peter Albert, Amjad Almahairi, Yasmine Babaei, Nikolay Bashlykov, Soumya Batra, Prajjwal Bhargava, Shruti Bhosale, et~al. 2023.
\newblock Llama 2: Open foundation and fine-tuned chat models.
\newblock \emph{arXiv preprint arXiv:2307.09288}.

\bibitem[{Wan et~al.(2024)Wan, Huang, Cai, Quan, Bi, and Shi}]{wan2024knowledge}
Fanqi Wan, Xinting Huang, Deng Cai, Xiaojun Quan, Wei Bi, and Shuming Shi. 2024.
\newblock Knowledge fusion of large language models.
\newblock \emph{arXiv preprint arXiv:2401.10491}.

\bibitem[{Wang et~al.(2020)Wang, Cho, and Gu}]{wang2020neural}
Changhan Wang, Kyunghyun Cho, and Jiatao Gu. 2020.
\newblock Neural machine translation with byte-level subwords.
\newblock In \emph{Proceedings of the AAAI conference on artificial intelligence}, volume~34, pages 9154--9160.

\bibitem[{Wang et~al.(2023{\natexlab{a}})Wang, Cheng, Zhan, Li, Song, and Liu}]{wang2023openchat}
Guan Wang, Sijie Cheng, Xianyuan Zhan, Xiangang Li, Sen Song, and Yang Liu. 2023{\natexlab{a}}.
\newblock Openchat: Advancing open-source language models with mixed-quality data.
\newblock In \emph{The Twelfth International Conference on Learning Representations}.

\bibitem[{Wang et~al.(2023{\natexlab{b}})Wang, Polo, Sun, Kundu, Xing, and Yurochkin}]{wang2023fusing}
Hongyi Wang, Felipe~Maia Polo, Yuekai Sun, Souvik Kundu, Eric Xing, and Mikhail Yurochkin. 2023{\natexlab{b}}.
\newblock Fusing models with complementary expertise.
\newblock In \emph{Annual Conference on Neural Information Processing Systems}.

\bibitem[{Wang et~al.(2022)Wang, Xie, Li, Fan, Song, Liang, Lu, Luo, and Shao}]{wang2021pvtv2}
Wenhai Wang, Enze Xie, Xiang Li, Deng-Ping Fan, Kaitao Song, Ding Liang, Tong Lu, Ping Luo, and Ling Shao. 2022.
\newblock Pvtv2: Improved baselines with pyramid vision transformer.
\newblock \emph{Computational Visual Media}, 8(3):1--10.

\bibitem[{Wang et~al.(2017)Wang, Luo, Crankshaw, Tumanov, Yu, and Gonzalez}]{wang2017idk}
Xin Wang, Yujia Luo, Daniel Crankshaw, Alexey Tumanov, Fisher Yu, and Joseph~E Gonzalez. 2017.
\newblock Idk cascades: Fast deep learning by learning not to overthink.
\newblock \emph{arXiv preprint arXiv:1706.00885}.

\bibitem[{Xu et~al.(2024)Xu, Lu, and Zhang}]{xu2024bridging}
Yangyifan Xu, Jinliang Lu, and Jiajun Zhang. 2024.
\newblock Bridging the gap between different vocabularies for llm ensemble.
\newblock \emph{arXiv preprint arXiv:2404.09492}.

\bibitem[{Yadav et~al.(2024)Yadav, Tam, Choshen, Raffel, and Bansal}]{yadav2024ties}
Prateek Yadav, Derek Tam, Leshem Choshen, Colin~A Raffel, and Mohit Bansal. 2024.
\newblock Ties-merging: Resolving interference when merging models.
\newblock \emph{Advances in Neural Information Processing Systems}, 36.

\bibitem[{Young et~al.(2024)Young, Chen, Li, Huang, Zhang, Zhang, Li, Zhu, Chen, Chang et~al.}]{young2024yi}
Alex Young, Bei Chen, Chao Li, Chengen Huang, Ge~Zhang, Guanwei Zhang, Heng Li, Jiangcheng Zhu, Jianqun Chen, Jing Chang, et~al. 2024.
\newblock Yi: Open foundation models by 01. ai.
\newblock \emph{arXiv preprint arXiv:2403.04652}.

\bibitem[{Yu et~al.(2023)Yu, Yu, Yu, Huang, and Li}]{yu2023language}
Le~Yu, Bowen Yu, Haiyang Yu, Fei Huang, and Yongbin Li. 2023.
\newblock Language models are super mario: Absorbing abilities from homologous models as a free lunch.
\newblock \emph{arXiv preprint arXiv:2311.03099}.

\bibitem[{Zhang et~al.(2023)Zhang, Press, Merrill, Liu, and Smith}]{zhang2023language}
Muru Zhang, Ofir Press, William Merrill, Alisa Liu, and Noah~A Smith. 2023.
\newblock How language model hallucinations can snowball.
\newblock \emph{arXiv preprint arXiv:2305.13534}.

\bibitem[{Zheng et~al.(2023)Zheng, Chiang, Sheng, Zhuang, Wu, Zhuang, Lin, Li, Li, Xing, Zhang, Gonzalez, and Stoica}]{zheng2023judging}
Lianmin Zheng, Wei-Lin Chiang, Ying Sheng, Siyuan Zhuang, Zhanghao Wu, Yonghao Zhuang, Zi~Lin, Zhuohan Li, Dacheng Li, Eric.~P Xing, Hao Zhang, Joseph~E. Gonzalez, and Ion Stoica. 2023.
\newblock \href {https://arxiv.org/abs/2306.05685} {Judging llm-as-a-judge with mt-bench and chatbot arena}.
\newblock \emph{Preprint}, arXiv:2306.05685.

\end{thebibliography}

\appendix
\newpage
\section{Step-by-Step Examples of \textsc{GaC}}
\label{sec:appendix-example}

To better demonstrate the \textsc{GaC} ensemble, we posed two classic and interesting questions to two individual models (OpenChat-3.5-0106 and SOLAR-10.7B-Instruct-v1.0),  as well as to their ensemble. 
\begin{itemize}
    \item \it Q1: Which word does not belong with the others?\textbackslash n tyre, steering wheel, car, engine
    \item \it Q2: 9.11 and 9.9, which is bigger?
\end{itemize}
Their responses are listed in Tab.\ref{tab:Q1Q2_responses}. Next, we list the tokens generated by the two models at each step during the ensemble process that have the highest confidence, as well as the token chosen by the \textsc{GaC}, in Tab.\ref{tab:ensemble_step_by_step}.

In Q1, OpenChat tried to output incorrect answers "ste" and "ty" directly in step 1 and step 4, but they were not adopted due to insufficient confidence. In step 4, SOLAR outputted the high confidence token "car", which resulted in the ensemble answer being correct.

\begin{table*}[ht]
\footnotesize
\centering
\begin{tabular}{p{0.24\textwidth} p{0.335\textwidth} p{0.335\textwidth}}
\toprule
 & \textbf{Response of Q1} & \textbf{Response of Q2} \\
\midrule
OpenChat-3.5-0106 & 
\it steering wheel. & 
\it 9.11 is bigger than 9.9. \\
\midrule

SOLAR-10.7B-Instruct-v1.0 & 
\it The word "car" does not belong with the others as the other options are components or parts of a car, while "car" is the overall vehicle itself. & 
\it To determine which number is bigger, we compare the numbers digit by digit from the rightmost digit (the decimal point) to the left. In this case, we have 1 in the decimal place for 9.11 and 9 in the decimal place for 9.9. Since 9 is greater than 1, we can conclude that 9.9 is the bigger number.\\
\midrule

\textsc{GaC} (OpenChat + SOLAR) & 
\it The word "car" does not belong with the others. The other words (tyre, steering wheel, and engine) are all parts of a car, while "car" is the whole vehicle itself. & 
\it 9.9 is bigger than 9.11. However, it seems like there might be a mistake in the numbers provided. The number 9.11 is commonly associated with the date of the September 11, 2001 terrorist attacks in the United States, and it is not a typical number to compare with 9.9. If you meant to compare 9.1 and 9.9, then 9.9 is indeed bigger. \\
\midrule

Ground truth & 
\it Car does not belong because all others are components of a car. & 
\it 9.9 \\
\bottomrule
\end{tabular}
\caption{Responses and groud truth of Q1 and Q2}
\label{tab:Q1Q2_responses}
\end{table*}

\begin{table*}[!ht]
\footnotesize
\centering
\begin{tabular}{>{\centering\arraybackslash}p{0.05\textwidth} 
                                   >{\raggedright\arraybackslash}p{0.38\textwidth} 
                                   >{\raggedright\arraybackslash}p{0.39\textwidth}}
\toprule
\textbf{Step} & \textbf{Q1 Ensemble (OpenChat / SOLAR / \textsc{GaC})} & \textbf{Q2 Ensemble (OpenChat / SOLAR / \textsc{GaC})} \\
\midrule
1 & 
\it ste (0.419) / The (0.950) / The & 
\it \tokenSpace ~(0.976) / To (0.603) / \tokenSpace \\
\midrule

2 & 
\it \tokenSpace word (0.994) / \tokenSpace word (1.000) / \tokenSpace word & 
\it 9 (1.000) / \textbackslash n (1.000) / \textbackslash n \\
\midrule

3 & 
\it \tokenSpace that (0.594) / \tokenSpace " (0.999) / \tokenSpace " & 
\it 9 (0.942) / When (0.342) / 9 \\
\midrule

4 & 
\it ty (0.714) / car (1.000) / car & 
\it . (1.000) / . (1.000) / . \\
\midrule

5 & 
\it " (0.999) / " (1.000) / " & 
\it 1 (0.689) / 9 (1.000) / 9 \\
\midrule

6 & 
\it \tokenSpace does (0.992) / \tokenSpace does (0.998) / \tokenSpace does & 
\it \tokenSpace is (0.994) / \tokenSpace is (1.000) / \tokenSpace is \\
\midrule

7 & 
\it \tokenSpace not (1.000) / \tokenSpace not (1.000) / \tokenSpace not & 
\it \tokenSpace bigger (0.969) / \tokenSpace a (0.765) / \tokenSpace bigger \\
\midrule

8 & 
\it \tokenSpace belong (0.999) / \tokenSpace belong (0.950) / \tokenSpace belong & 
\it \tokenSpace than (0.645) / \tokenSpace than (1.000) / \tokenSpace than \\
\midrule

9 & 
\it \tokenSpace with (0.999) / \tokenSpace with (0.990) / \tokenSpace with & 
\it \tokenSpace  ~(1.000) / \tokenSpace  ~(1.000) / \tokenSpace \\
\midrule

10 & 
\it \tokenSpace the (1.000) / \tokenSpace the (1.000) / \tokenSpace the & 
\it 9 (1.000) / 9 (1.000) / 9 \\
\midrule

11 & 
\it \tokenSpace others (0.889) / \tokenSpace others (0.990) / \tokenSpace others & 
\it . (1.000) / . (1.000) / . \\
\midrule

12 & 
\it . (0.726) / as (0.638) / . & 
\it 1 (1.000) / 1 (1.000) / 1 \\
\midrule

13 & 
\it \tokenSpace The (0.428) / \tokenSpace The (0.942) / \tokenSpace The & 
\it 1 (1.000) / 1 (1.000) / 1 \\
\midrule

14 & 
\it \tokenSpace other (0.866) / \tokenSpace other (0.982) / \tokenSpace other & 
\it . (0.948) / . (0.483) / . \\
\midrule
... & 
... & 
... \\
% ...
\bottomrule
\end{tabular}
\caption{The highest confidence token in each step of individual models (confidence indicated in parentheses) and the token chosen by \textsc{GaC}.}
\label{tab:ensemble_step_by_step}
\end{table*}

\section{Benchmarks}
\label{sec:appendixA}

This paper uses the lm-evaluation-harness v0.4.1. The task names in the repo corresponding to each of the benchmarks we used are as follows:

\noindent MMLU: \emph{mmlu\_flan\_n\_shot\_generative}, 5-shots

\noindent GSM8K: \emph{gsm8k}, 5-shots.

\noindent BBH: \emph{bbh\_fewshot}, 3-shots.

\noindent TriviaQA: \emph{triviaqa}, 5-shots.

\noindent NQ: \emph{nq\_open}, 5-shots.

\section{Hardware Specifications}
\label{sec:appendixB}

We list the hardware corresponding to the models used in this paper in Tab.\ref{tab:models-hardware}.

\begin{table*}[!ht]
    \centering
    \footnotesize
    \begin{tabular}{@{}>{\raggedright\arraybackslash}p{4.7cm} p{1.5cm}@{}}
        \toprule
        \textbf{Models} & \textbf{Hardware} \\
        \midrule
        Phi-3-mini-4k-instruct & 1x A100 \\
        Llama-3-8B-Instruct & 1x A100 \\
        openchat\_3.5 & 1x A100 \\
        OpenChat-3.5-0106 & 1x A100 \\
        Qwen1.5-14B-Chat & 1x A100 \\
        SOLAR-10.7B-Instruct-v1.0 & 1x A100 \\
        Nous-Hermes-2-SOLAR-10.7B & 1x A100 \\
        Yi-34B-Chat & 1x A100 \\
        Qwen1.5-32B-Chat & 1x A100 \\
        Nous-Hermes-2-Mixtral-8x7B-DPO & 2x A100 \\
        Mixtral-8x7B-Instruct-v0.1 & 2x A100 \\
        Llama-3-70B-Instruct & 2x A100 \\
        Qwen1.5-72B-Chat & 3x A100 \\
        Qwen2-72B-Instruct & 3x A100 \\
        \bottomrule
    \end{tabular}
    \caption{Models and Hardware}
    \label{tab:models-hardware}
\end{table*}

\section{Translation and Summarization}
\label{sec:appendix-TS}

We applied the \textsc{GaC} ensemble to OpenChat-3.5-0106 and SOLAR-10.7B-Instruct-v1.0 to evaluate its performance on translation and summarization tasks \citep{bojar-EtAl:2014:W14-33, bojar-EtAl:2016:WMT1,narayan2018don}. The results are presented in Tab.\ref{tab:gac_TS}.

\begin{table*}[ht]
\footnotesize
\centering
\begin{tabular}{@{}lccccccccc@{}}
    \toprule
    \textbf{Models} & \multicolumn{2}{c}{\textbf{EN → FR}} & \multicolumn{2}{c}{\textbf{FR → EN}} & \multicolumn{2}{c}{\textbf{EN → DE}} & \multicolumn{2}{c}{\textbf{DE → EN}} & \textbf{XSUM} \\
    \cmidrule(lr){2-3} \cmidrule(lr){4-5} \cmidrule(lr){6-7} \cmidrule(lr){8-9} \cmidrule(lr){10-10}
    & \textbf{BLEU} & \textbf{CHRF} & \textbf{BLEU} & \textbf{CHRF} & \textbf{BLEU} & \textbf{CHRF} & \textbf{BLEU} & \textbf{CHRF} & \textbf{ROUGE-L} \\
    \midrule
    OpenChat-3.5-0106 & 35.40 & 61.71 & 40.52 & 64.82 & 31.48 & 59.84 & 44.32 & 67.97 & 26.54 \\
    SOLAR-10.7B-Instruct-v1.0 & 35.61 & 62.15 & 40.28 & 64.86 & 30.90 & 60.06 & 42.85 & 67.79 & 23.81 \\
    \textsc{GaC} (OpenChat + SOLAR) & 36.59 & 62.82 & 41.63 & 65.56 & 33.10 & 61.43 & 44.65 & 68.69 & 27.26 \\
    \midrule
    $\Delta$ & +0.98 & +0.67 & +1.11 & +0.70 & +1.62 & +1.37 & +0.33 & +0.72 & +0.72 \\
    \bottomrule
\end{tabular}
\caption{Performance on translation and summarization tasks. $\Delta$ represents the improvement compared to the best individual model.}
\label{tab:gac_TS}
\end{table*}

\end{document}